\documentclass[lettersize,journal]{IEEEtran}
\usepackage{amsmath,amsfonts}
\usepackage{algorithmic}
\usepackage{algorithm}
\usepackage{array}
\usepackage[caption=false,font=normalsize,labelfont=sf,textfont=sf]{subfig}
\usepackage{textcomp}
\usepackage{stfloats}
\usepackage{url}
\usepackage{verbatim}
\usepackage{graphicx}
\usepackage[
        numbers,square,
        ]{natbib}

\usepackage{xcolor}
\usepackage{graphicx}
\usepackage{relsize}
\definecolor{cvprblue}{HTML}{000000}
\definecolor{r2color}{HTML}{000000}
\usepackage[
        breaklinks,
        colorlinks,
        citecolor=blue,
        linkcolor=blue,
        ]{hyperref}

\usepackage{multirow}
\usepackage{array}
\begin{document}

\title{Towards Full-scene Domain Generalization in Multi-agent Collaborative Bird's Eye View Segmentation for Connected and Autonomous Driving}

\author{Senkang Hu, Zhengru Fang,~\IEEEmembership{Graduate Student Member,~IEEE,} Yiqin Deng, Xianhao Chen,~\IEEEmembership{Member,~IEEE,} \\ Yuguang Fang,~\IEEEmembership{Fellow,~IEEE,} Sam Kwong,~\IEEEmembership{Fellow,~IEEE}

\thanks{This work was supported in part by the Hong Kong Innovation and Technology Commission under InnoHK Project CIMDA, in part by the Hong Kong SAR Government under the Global STEM Professorship and Research Talent Hub, and in part by the Hong Kong Jockey Club under the Hong Kong JC STEM Lab of Smart City (Ref.: 2023-0108). The work of Yiqin Deng was supported in part by the National Natural Science Foundation of China under Grant No. 62301300. The work of Xianhao Chen was supported in part by HKU-SCF FinTech Academy R\&D Funding. \textit{(Corresponding Author: Yiqin Deng)}}
\thanks{S. Hu, Z. Fang, Y. Deng and Y. Fang are with the Department of Computer
Science, City University of Hong Kong, Hong Kong. (e-mail: \texttt{\{senkang.forest, zhefang4-c\}@my.cityu.edu.hk, \{yiqideng, my.Fang\}@cityu.edu.hk})}
\thanks{X. Chen is with the Department of Electrical and Electronic Engineering,
The University of Hong Kong, Hong Kong. (e-mail: \texttt{xchen@eee.hku.hk})}
\thanks{S. Kwong is with the Department of Computing and Decision Sciences, Lingnan University, Hong Kong. (e-mail: \texttt{samkwong@ln.edu.hk})}
}


\maketitle
\begin{abstract}
Collaborative perception has recently gained significant attention in autonomous driving, improving perception quality by enabling the exchange of additional information among vehicles. However, deploying collaborative perception systems can lead to domain shifts due to diverse environmental conditions and data heterogeneity among connected and autonomous vehicles (CAVs). To address these challenges, we propose a unified domain generalization framework to be utilized during the training and inference stages of collaborative perception. In the training phase, we introduce an Amplitude Augmentation (AmpAug) method to augment low-frequency image variations, broadening the model's ability to learn across multiple domains. We also employ a meta-consistency training scheme to simulate domain shifts, optimizing the model with a carefully designed consistency loss to acquire domain-invariant representations. In the inference phase, we introduce an intra-system domain alignment mechanism to reduce or potentially eliminate the domain discrepancy among CAVs prior to inference. Extensive experiments substantiate the effectiveness of our method in comparison with the existing state-of-the-art works. 
\end{abstract}

\begin{IEEEkeywords}
Domain generalization, vehicle-to-vehicle collaborative perception, autonomous driving, bird's eye view segmentation.
\end{IEEEkeywords}

\section{Introduction}

\begin{figure}[t]
    \centering
     \includegraphics[width=.7\linewidth]{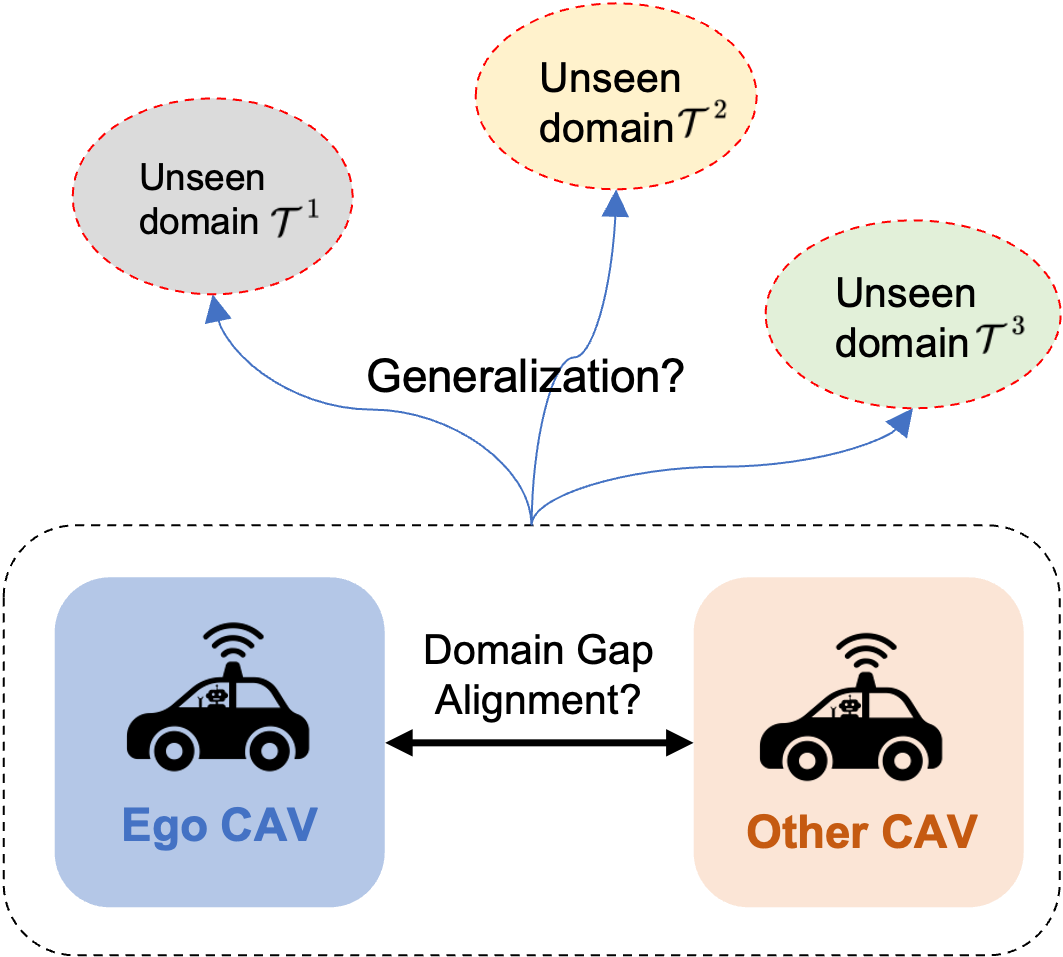}
  
    \caption{\textcolor{cvprblue}{\textbf{The problem setting} of domain generalization for collaborative perception, which aims to tackle the domain generalization problem in collaborative perception while aligning the domain gap among different CAVs.}}
    \label{fig:problem_illustration}
    \vspace{-3mm}
\end{figure}

\IEEEPARstart{R}{ecently}, multi-agent collaborative perception \cite{huWhere2commCommunicationefficientCollaborative2024,liuWhen2comMultiAgentPerception2020,yangWhat2commCommunicationefficientCollaborative2023,chenVehicleServiceVaaS2024,huAgentsCoDriverLargeLanguage2024,xuV2XViTVehicletoEverythingCooperative2022a} has attracted increasing attention in the autonomous driving community, due to its promising way to overcome the limitations of a single-agent perception system. For example, a single-agent perception system may suffer from occlusion and sparse sensor observation from afar, which may lead to a high risk of traffic accidents. Collaborative perception can address these issues by employing vehicle-to-vehicle (V2V) collaboration, where visual information (e.g., raw sensory information, intermediate perception features, and perception results) from multiple nearby CAVs can be shared to improve the accurate understanding of the \textcolor{r2color}{environments}.


In collaborative perception, current approaches mainly aim to strike an \textcolor{r2color}{optimal} balance between performance and bandwidth consumption \textcolor{r2color}{through} developing new perception architectures. For example, Liu \textit{et al.} \cite{liuWho2comCollaborativePerception2020} leveraged a three-step handshake communication protocol to determine the information with which CAVs should be shared. Chen \textit{et al.} \cite{chenCooperCooperativePerception2019} conducted a study to implement collaborative perception by fusing raw LiDAR point clouds of different CAVs. 
Xu \textit{et al.} \cite{xuV2XViTVehicletoEverythingCooperative2022a} proposed a new vision transformer framework to \textcolor{r2color}{fuse the information from the on-road vehicles and the roadside units (RSU)}. Hu \textit{et al.} \cite{huAdaptiveCommunicationsCollaborative2023GLOBECOM} proposed an adaptive communication scheme that can construct the communication graph and minimize the communication delay according to different channel information state (CSI). 

\textcolor{r2color}{Despite the aforementioned methods} having made significant progress in balancing performance and bandwidth while optimizing communication graph construction, they regrettably overlook a critical design challenge\textcolor{r2color}{, namely, the domain shift problem in collaborative perception}. These methods typically assume that both training and testing data originate from the same domain, a condition rarely met in real-world applications.

The consequence of neglecting the domain shift is a severe deterioration in perception performance when these methods are implemented in environments characterized by domain variations. For instance, in OPV2V \cite{xuOPV2VOpenBenchmark2022} and nuScenes \cite{caesarNuScenesMultimodalDataset2020}, the widely recognized benchmarks for bird's eye view (BEV) semantic segmentation, their data collection settings exhibit notable disparities, encompassing differences in intrinsic and extrinsic parameters. One dataset is sourced from a simulator, while the other derives from real-world observations. When a model initially trained on OPV2V is directly transferred to the nuScenes dataset, a substantial performance drop is observed. Consequently, there is an urgent need to achieve domain generalization for collaborative perception.

Domain generalization (DG) is an active research topic \textcolor{r2color}{that has been investigated by many researchers before} \cite{liDomainAdaptationBased2024,zhouDomainGeneralizationMixStyle2021,liDomainGeneralizationAdversarial2018,yangFDAFourierDomain2020,volpiGeneralizingUnseenDomains2018}.
 However, when applied to collaborative perception in the context of CAVs, the domain generalization problem becomes notably more complicated. Several challenges make collaborative perception in CAVs much more demanding for DG. First, for collaborative BEV segmentation, there are few open-source benchmarks, and annotating data is resource-intensive and time-consuming. 
Only OPV2V \cite{xuOPV2VOpenBenchmark2022} and V2X-Sim \cite{liV2XSimMultiAgentCollaborative2022} provide accessible datasets. Moreover, 
\textcolor{r2color}{the environments of CAVs are highly dynamic,} subject to frequent changes. For example, domains will shift when transitioning from city roads to country roads, and similar domain shifts will also be observed when moving from daytime to night conditions.
Furthermore, \textcolor{r2color}{data heterogeneity among CAVs} also presents a significant challenge. In practical scenarios, different \textcolor{r2color}{CAVs} may operate under diverse environmental conditions during collaborative perception. For instance, one \textcolor{r2color}{CAV} might navigate in darkness, while another is positioned in a well-lit area. The absence of standardized parameters across different onboard cameras lead to variations in image quality, affecting factors like brightness, contrast, and color \cite{zhouDomainGeneralizationSurvey2022}. As a result, these divergent environmental conditions and sensor characteristics inevitably create a substantial domain gaps among \textcolor{r2color}{CAVs}, resulting in performance degradation in collaborative perception.

To tackle these challenges, \textcolor{r2color}{we present a comprehensive domain generalization framework that operates seamlessly in both training and inference phases of collaborative perception}. Firstly, our approach starts with the development of an amplitude augmentation method, a resource-efficient technique that does not need extra annotated data and is ready for immediate use. Next, we introduce a meta-consistency training scheme rooted in meta-learning principle. This scheme equips the model with the ability to generalize effectively to unseen domains while fostering the acquisition of domain-invariant features. Finally, our framework includes an intra-system domain alignment scheme, strategically designed to mitigate data heterogeneity and bridge the domain gaps among various CAVs. The main contributions of this paper are summarized as follows.
\begin{itemize}
    \item We address the practical problem of domain generalization for collaborative perception. \textcolor{r2color}{It is the first work to study domain generalization in collaborative BEV perception.}
    \item We delve into the underlying reasons for domain shifts at the image level and introduce an amplitude augmentation method to enhance the low-level distribution of images. Additionally, we collect a dataset that aids in augmenting the source domain for application to different target domains.
    \item We establish a meta-consistency training scheme based on meta-learning which guides the model to adapt to new domains. We also carefully design a consistency loss to constrain the feature distribution between the features during meta-training and meta-testing phases, which can encourage the model to learn the domain-invariant representations. 
    \item We devise an intra-system domain alignment scheme tailored to reduce data heterogeneity and bridge domain gaps among diverse CAVs.
\end{itemize}

The remainder of this paper is organized as follows. In Sec. \ref{sec:related_work}, we review the related work of collaborative perception and domain generalization. In Sec. \ref{sec:method}, we introduce our proposed domain generalization method in detail. In Sec. \ref{sec:experiments}, we conduct comprehensive experiments to evaluate our method. Finally, we conclude our work in Sec. \ref{sec:conclusion}.

\section{Related Work}
\label{sec:related_work}

\subsection{Collaborative Perception (CP)}

Despite the notable progress in autonomous driving in recent years, single-agent perception systems face significant challenges with respect to occlusions and limitations of sensor coverage. For example, in a crowded urban environment, a vehicle may be blocked by other vehicles, pedestrians, or buildings, which make it difficult to perceive its surrounding {environments}.
Multi-agent collaborative perception (CP) emerged as a solution to addressing these issues \cite{wangV2VNetVehicletoVehicleCommunication2020a,liuWho2comCollaborativePerception2020,fang2024pacp,huCollaborativePerceptionConnected2024}. 

Based on data sharing strategies, CP can generally be divided into three aspects: 1) early fusion \cite{chenCooperCooperativePerception2019}, in which raw data is shared with ego CAV; 2) intermediate fusion, in which intermediate features extracted from each CAV's sensor data are shared; and 3) late fusion, in which perception outputs are shared (e.g., bounding box position, segmentation mask, etc.). Recent works indicate that intermediate fusion is more efficient than early fusion and late fusion because it effectively  balances accuracy and communication bandwidth. For example,
Liu \textit{et al.} \cite{liuWho2comCollaborativePerception2020} proposed a handshake communication mechanism, which consists of three stages, namely request, match, and connect, determining which \textcolor{r2color}{CAVs} should communicate with. Li \textit{et al.} \cite{liLearningDistilledCollaboration2021} leveraged knowledge distillation to train a model by a teacher-student framework and proposed a design of a matrix to allow an \textcolor{r2color}{CAV} to adaptively highlight the information region, thereby achieving effective intermediate fusion. Wang \textit{et al.} \cite{wangV2VNetVehicletoVehicleCommunication2020a} proposed an intermediate fusion strategy where all \textcolor{r2color}{CAVs} transmit features derived from the raw point cloud to strike a balance between bandwidth and precision. Liu \textit{et al.} proposed CRCNet \cite{liuCRCNetFewshotSegmentation2022}, a few-shot segmentation network that leveraged cross-reference and local-global condition to better discover the concurrent objects in two images to enhance the few-shot segmentation learning. Wang \textit{et al.} \cite{wangCollaborativeVisualNavigation2021} proposed a 3D dataset, CollaVN, to facilitate the development of multi-agent collaborative visual navigation. Moreover, Su \textit{et al.} \cite{suUncertaintyQuantificationCollaborative2023} estimated the uncertainty of the object detection in CP and Hu \textit{et al.} \cite{huCollaborationHelpsCamera2023} presented Coca3D to enhance camera-only 3D detection through the incorporation of multi-agent collaborations.
In addition to the methods mentioned above, numerous datasets have also emerged in the field of CP to facilitate the development of this field, such as OPV2V \cite{xuOPV2VOpenBenchmark2022}, V2X-Sim \cite{liV2XSimMultiAgentCollaborative2022}, V2XSet \cite{xuV2XViTVehicletoEverythingCooperative2022a}, DAIR-V2X \cite{yuDAIRV2XLargeScaleDataset2022}, and V2V4Real \cite{xuV2V4RealRealWorldLargeScale2023}.
Despite the aforementioned advances of fusion methods, the domain shift caused by different environmental conditions and data heterogeneity among CAVs has not been well investigated in CP, which is the topic of this paper.   

\subsection{Domain Generalization}

The domain shift problem has seriously hindered large-scale deployments of machine learning models. To tackle this issue, a lot of DG methods have been proposed \cite{gongDLOWDomainFlow2021,ghifaryDomainGeneralizationObject2015,liEpisodicTrainingDomain2019,yangFDAFourierDomain2020,liFeatureCriticNetworksHeterogeneous2019}. DG is dedicated to learning a model trained on multiple source domains that can generalize to unseen target domains. For example, Motiian \textit{et al.} \cite{motiianUnifiedDeepSupervised2017} leveraged a contrastive loss to learn an embedding subspace that is discriminative, and the mapped visual domains are semantically aligned and yet maximally separated, which requires only a few labeled target samples. Some other \textcolor{r2color}{DG} methods are based on data augmentation, which is a common practice to regularize the training of machine learning models to avoid overfitting and improve generalization, e.g., Volpi \textit{et al.} \cite{volpiAddressingModelVulnerability2019} defined a new data augmentation rule to transform images with different distribution where the model is most vulnerable, thereby making the model more robust against distribution shifts. Volpi \textit{et al.} \cite{volpiGeneralizingUnseenDomains2018} also proposed a data augmentation method to leverage adversarial gradients obtained from a task classifier to perturb input images. In addition, meta-learning, also known as learn-to-learn, is another method helpful for \textcolor{r2color}{DG}, which exposes a model to domain shift and guides the model to learn how to learn from the domain shift.
Finn \textit{et al.} \cite{finnModelAgnosticMetaLearningFast2017} proposed MAML, a method that partitions the training data into meta-train and meta-test subsets, and trains a model using the meta-train subset to enhance its performance on the meta-test subset. Unfortunately, most aforementioned works are not specifically on CP. 
Our method also utilizes meta-learning scheme as our foundational learning scheme for CP. Here, we propose a novel amplitude augmentation method to simulate the domain shift, and leverage a consistency optimization to enable the model to learn domain-invariant features.

\begin{figure*}[t]
    \centering
    \includegraphics[width=1\textwidth]{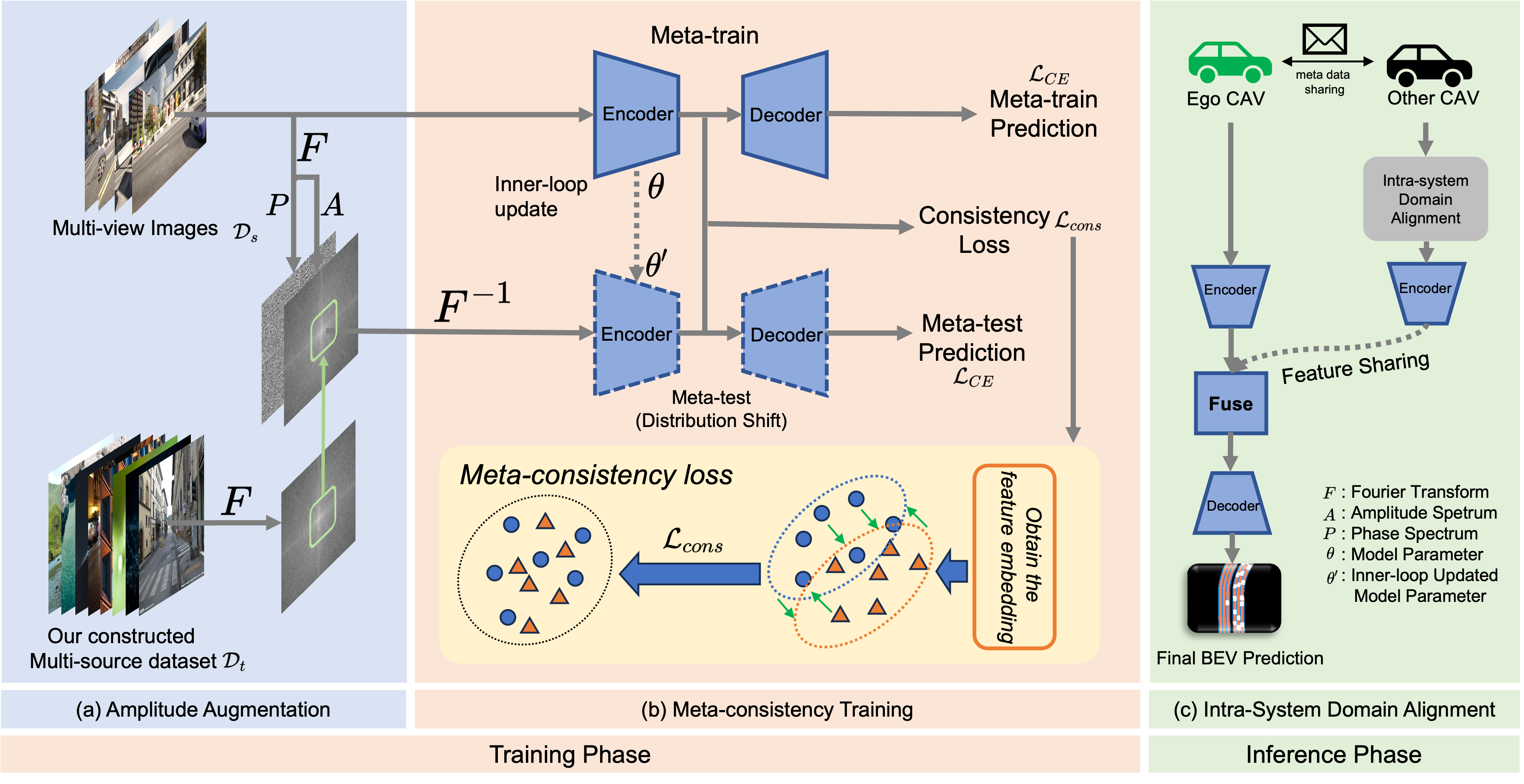}
    \caption{\textbf{Overall Architecture.} 1) The first part of our proposed method is amplitude augmentation which can transform the original source domain
    data to different target domains. 2) The second part is meta-consistency training, which can simulate the domain shift, guiding the model to learn how to learn from different domains. Then, we exploit the meta-consistency loss to encourage our model to learn the domain-invariant features, thereby enhancing the ability of generalization of the model. 3) The third part is the intra-system domain alignment, which can minimize the domain gap among the
    data perceived by different collaborative vehicles prior to
    inference.}
    \label{fig:overall_architecture}
    \vspace{-3mm}
\end{figure*}
\section{Our Method}
\label{sec:method}

\subsection{Problem Formulation and Overview}

\textbf{Collaborative Domain Generalization.}
Let $(\mathcal{X}, \mathcal{Y})$ denote the input and label space in a collaborative BEV segmentation task, $\mathcal{D} = \{\mathcal{D}_1, \mathcal{D}_2, \cdots, \mathcal{D}_K\}$ be the set of $K$ source domains. In each domain, $\mathcal{D}_k=\{(x_i^{(k)}, y_i^{(k)} )\}^{N_k}_{i=1}$ is a set of samples from domain distribution ($\mathcal{X}_k, \mathcal{Y}$), where $x_i^{(k)}$ is the input image set of CAVs at a certain time, $y_i^{(k)}$ is the corresponding label, and $N_k$ is the number of samples in domain $k$. The goal is to learn a collaborative BEV segmentation model $F: \mathcal{X} \rightarrow \mathcal{Y}$ based on the data from multiple source domains $\mathcal{D}$, which can be generalized to the unseen target domain.

\textbf{Overview.} Our approach offers a unified domain generalization framework for both training and inference in collaborative perception, making it broadly applicable to encoder-decoder structures.
Specifically, we first propose an \textit{amplitude augmentation} (AmpAug) method, which enhances the model's robustness against domain discrepancies in CAV driving. AmpAug leverages fast Fourier transform to convert images into the frequency domain, incorporating low-frequency signals from our target dataset's amplitude spectrum. This process culminates in a new synthetic amplitude spectrum, seamlessly integrated with the source domain's phase spectrum. The result is an augmented image achieved through inverse Fourier transform.

Following AmpAug, we propose our \textit{meta-consistency training} scheme, a method capable of simulating domain shifts and optimizing the model using a carefully crafted consistency loss function. This loss drives the encoder to learn domain-invariant features by minimizing the maximum mean discrepancy of the features obtained during both meta-training and meta-testing phases.
Finally, our \textit{intra-system domain alignment} mechanism is introduced, \textcolor{r2color}{aimed at mitigating or even eliminating domain discrepancies among CAVs before the inference stage}. This is achieved by translating image styles and unifying the distribution of image pixels in color space. Each of these components will be discussed in detail in the following sections. The overall framework is shown in Fig. \ref{fig:overall_architecture}.

\subsection{Amplitude Augmentation}
\label{sec:amplitude_augmentation}
\begin{figure}[t]
    \centering
    \includegraphics[width=1\linewidth]{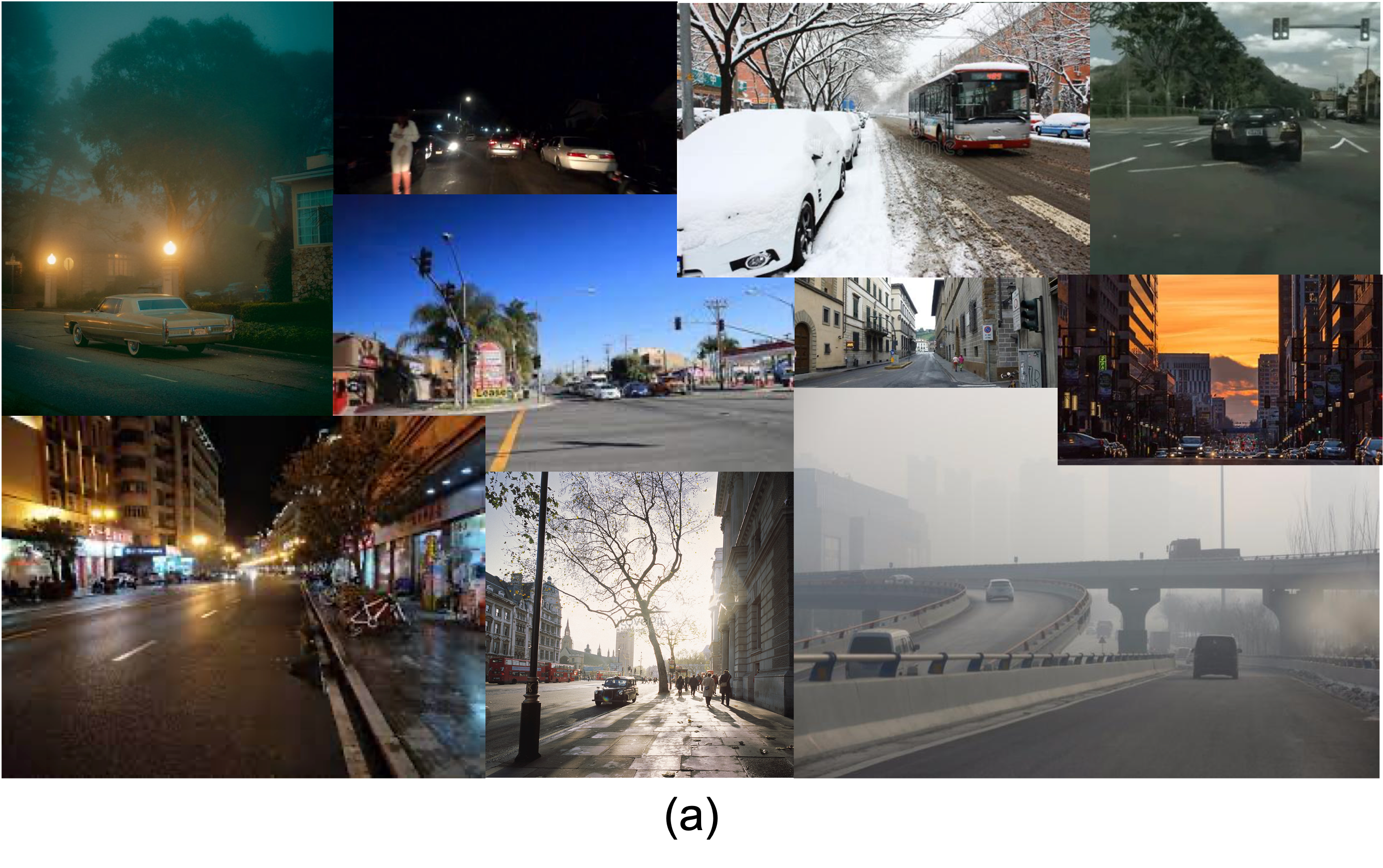}
    
    \includegraphics[width=1\linewidth]{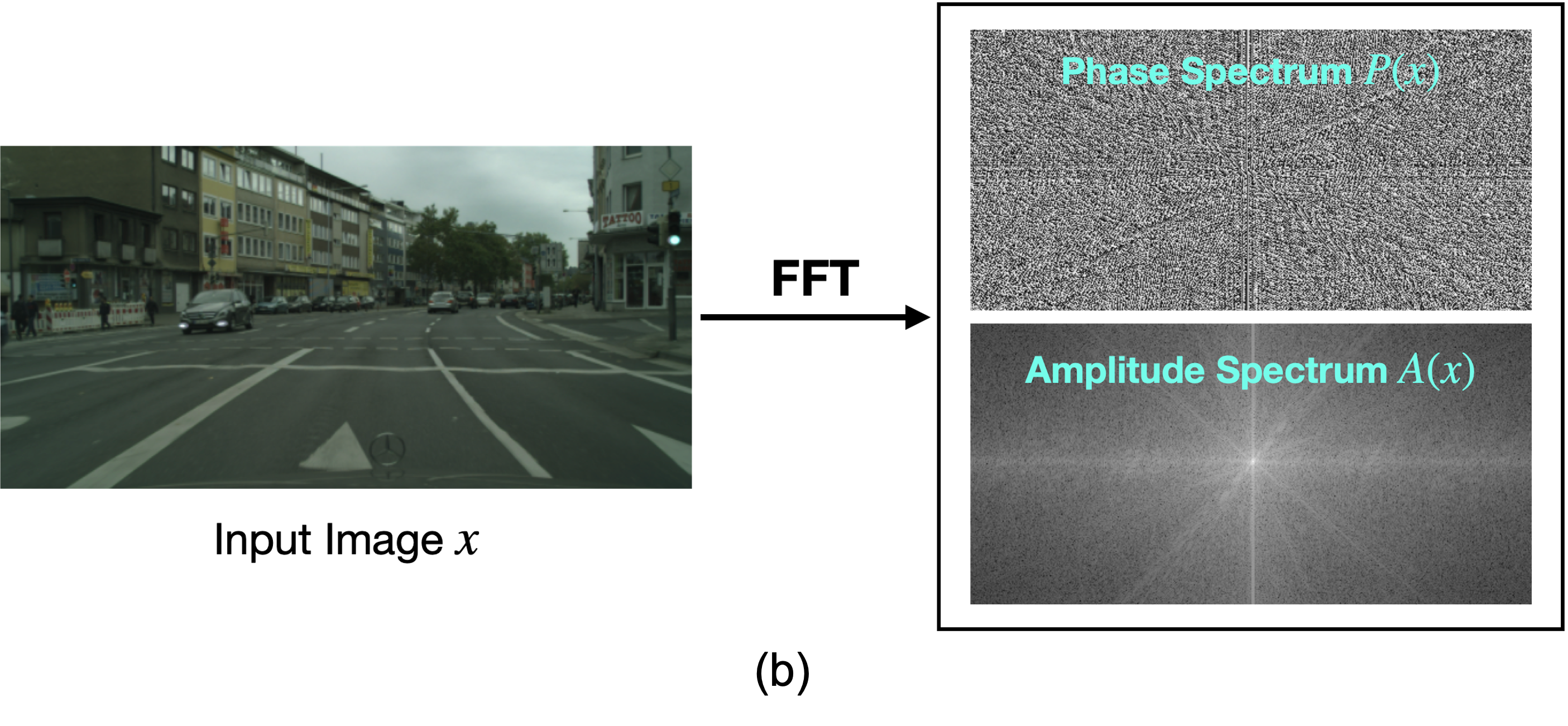}

    \caption{\textbf{Visualization of our constructed target dataset $\mathcal{D}_t$ and image fast Fourier transform (FFT).} \textcolor{cvprblue}{Subfigure (a) gives a brief illustration of the dataset, where we observe that this dataset contains images with different weather conditions, driving scenarios, colors, etc.; Subfigure (b) visualizes FFT of images, where we transform an input image $x$ to the frequency domain by FFT and obtain the amplitude spectrum $A(x)$ (low-frequency spectrum) and phase spectrum $P(x)$ (high-frequency spectrum). The amplitude spectrum indicates the magnitude of each frequency component present in the image, which is crucial for understanding the image's texture and style contents. The phase spectrum of an image specifies the phase or phase shifts of different spatial frequency components present in the image, providing detailed information about the spatial arrangement and positioning of features within the image.
    }}
    \label{fig:dataset_display}
    \vspace{-3mm}
\end{figure}

As for collaborative perception, the environments sensed by CAVs are extraordinarily complex due to dynamic transformations across time and space. Factors, such as varying weather conditions, diverse driving scenarios, fluctuating illuminations, and distinct image colors, contribute to this complexity. These environmental fluctuations result in domain discrepancy, significantly impeding perception performance when deploying models in real-world settings.

Our study, upon a close examination, has revealed that this domain discrepancy primarily arises from variations in the low-frequency spectrum of images, which can be obtained by fast Fourier transform (FFT). From the perspective of image processing, the amplitude spectrum can reflect the low-frequency distribution, such as the style, which can result in the domain gap.
Importantly, these low-frequency variations do not affect the image's finer details, encompassing high-frequency spectrum elements like objects, cars, people, and textures \cite{yangFDAFourierDomain2020,liuFedDGFederatedDomain2021}. Furthermore, these low-frequency spectrum variations hold substantial sway over the {images}. If these variances are not adequately represented in the training dataset, the models fail to generalize effectively. Consequently, we introduce a straightforward yet highly effective method known as \textit{AmpAug}, which serves the purpose of augmenting these low-frequency variations, compelling the model to acquire domain-invariant representations. This augmentation, in turn, reduces the model's susceptibility to the influence of low-frequency spectrum variations in images.

\textcolor{r2color}{Based on the above observations, we derive our novel design approach. }First, we construct a dataset $\mathcal{D}_t$, which is carefully collected from the Internet. We set some keywords to filter the images and obtain images under different weather conditions (including sunny, rainy, and foggy days), driving scenarios, colors, etc. The visualization of our constructed target dataset $\mathcal{D}_t$ is shown in Fig. \ref{fig:dataset_display}. 
The reason for constructing this dataset is that we want to obtain \textcolor{r2color}{a sufficient number of different amplitude spectra of the images and leverage them} to augment the amplitude spectra of the images in the source domain dataset $\mathcal{D}_s$.

Specifically, consider our constructed dataset $\mathcal{D}_t = \{x_i^t\}^{N_t}_{i=1}$ and a source domain dataset $\mathcal{D}_s = \{x_i^s, y_i^s\}^{N_s}_{i=1}$, where $x_i^t, x_i^s \in \mathbb{R}^{H\times W\times C}$ ($H$ and $W$ are the height and width of the image, respectively, and $C$ is the number of channels), $y_i^s \in \mathbb{R}^{H\times W\times C}$ is the corresponding label. We first sample an image $x_i^t$ from our constructed dataset $\mathcal{D}_t$, and convert it to the frequency domain by the fast Fourier transform (FFT):
\begin{equation}
    \mathbf{F}(x_i^t)(u,v,c)=
    \sum^{H-1}_{h=0}\sum^{W-1}_{w=0}x_i^t(h,w,c)e^{-2j\pi\left(\frac{h}{H}u+\frac{w}{W}v\right)}.
\end{equation}
After obtaining the frequency representation $\mathbf{F}(x_i^t)$, we can decompose it into amplitude spectrum $A(x_i^t) \in \mathbb{R}^{H\times W\times C}$ and phase spectrum $P(x_i^t)\in \mathbb{R}^{H\times W\times C}$, which represents the low-level distributions (e.g., style, light, etc.) and high-level distributions (e.g., details and objects, etc.), respectively \cite{yangFDAFourierDomain2020}. Then, we sample another image $x_j^s$ from the source dataset and perform the same operation to obtain the amplitude spectrum $A(x_j^s)$ and phase spectrum $P(x_j^s)$. After that, we can obtain the augmented image $x_j^{s\rightarrow t}$ by incorporating the low-level distribution $A(x_j^s)$ and $A(x_i^t)$. 

\textcolor{r2color}{Furthermore, a binary mask $M$ is introduced to control the proportion of the low-level distribution $A(x_j^s)$ and $A(x_i^t)$ in the augmented image $x_j^{s\rightarrow t}$}, $M(h,w)=\mathbf{1}_{h\times w}$ where $h\in[-\alpha H: \alpha H]$, $ w\in[-\alpha W: \alpha W]$, $\alpha\in (0,1)$. Then, we generate a new amplitude spectrum distribution as follows:
\begin{equation}
    \label{efi}
    A(x_j^{s\rightarrow t})= (I-M)\cdot A(x^s_j) + M\cdot A(x^t_i),
\end{equation}
where $x^s_j$ and $x^t_i$ are sampled from source domain dataset $\mathcal{D}_s$ and target domain dataset $\mathcal{D}_t$, while $I$ denotes the identity matrix. After obtaining the synthetic amplitude spectrum, we integrate it with the source domain phase spectrum to generate the augmented image by inverse Fourier transform $\mathbf{F}^{-1}$:
\begin{equation}
    \label{ex}
    x^{s\rightarrow t}_i= \mathbf{F}^{-1}\left(A(x_j^{s\rightarrow t}), P(x^s_i)\right).
\end{equation}

\textbf{Discussion.}
Here we elaborate the rationale behind the proposed method. Several factors contribute to this design, with the first being its independence of annotation. In those scenarios related to CAVs, acquiring labeled data is both costly and labor-intensive, often cited as a bottleneck in research and development, and posing challenges for both industry and academia, especially when accounting for time-varying weather conditions (e.g., sunny, rainy, foggy, snowy), color discrepancies, and diverse driving scenarios. Within the domain of collaborative BEV segmentation, open-source datasets are scarce. To our best knowledge, only two datasets exist \cite{xuOPV2VOpenBenchmark2022,liV2XSimMultiAgentCollaborative2022}, both of which are derived from {simulators} rather than real-world environments. The lack of real-world data underscores the urge need to develop methods that can operate effectively in the absence of extensive annotated datasets. Based on this observation and restriction, we propose AmpAug without using additional annotation. The second notable advantage of our proposed method is its plug-and-use nature. This feature is particularly advantageous as it allows for effortless integration into existing training pipelines without any{modifications}. This plug-and-use characteristic not only enhances the usability of the method but also significantly reduces the time and resources required for implementation.

In addition, here we also elaborate on the choice of $\alpha$ in Eq. (\ref{efi}) and the impact of selecting different proportions $\alpha$. As for the mask $M$, we have $M(h,w)=\mathbf{1}_{h\times w}$, where $h\in[-\alpha H: \alpha H]$, $ w\in[-\alpha W: \alpha W]$, where $\alpha\in (0,1)$ controls the size of the mask. From Eq. (\ref{efi}), we can see that the mask $M$ controls the proportion of the source and target images. When $\alpha = 0$, the image $x^{s\rightarrow t}$ remains identical to the original source image $x^s$. On the other hand, when $\alpha = 1.0$, the amplitude of $x^s$ is completely substituted with the amplitude of $x^t$. We have observed in our practice that as $\alpha$ progresses towards 1.0, the image $x^{s\rightarrow t}$ increasingly resembles the target image $x^t$. However, this transition also introduces noticeable artifacts in the image. Therefore, we set $\alpha = 0.01$ to balance the trade-off between the source and target images.

\subsection{Meta-consistency Training}

\textbf{{Meta-training.}} We use the gradient-based meta-learning algorithm as our foundational learning scheme, which can learn a generalizable model by simulating the real-world domain shift in the training phase. In our settings, the domain shift comes from the data generated from the frequency domain with amplitude augmentation. Specifically, in each iteration, we consider the raw camera source data $\mathcal{D}_s$ perceived by CAVs' cameras as {meta-training} data and its counterparts generated from frequency domain by AmpAug as {meta-testing} data $\mathcal{D}_{s\rightarrow t}$. The meta-learning can be divided into two stages. We first update the model parameters $\theta$ using the {meta-training} data $\mathcal{D}_s$ with cross entropy $\mathcal{L}_{CE}$:
\begin{equation}
    \mathcal{L}_{CE}=\frac{1}{N}\sum^{N-1}_{i=0}\left(y_i\log\hat{y}_i + (1-y_i)\log(1-\hat{y}_i)\right),
\end{equation}
where $y_i$ is the ground truth label and $\hat{y}_i$ is the predicted segmentation map. The model parameters $\theta$ are updated by the gradient descent:
\begin{equation}
    \label{eq:inner_loop}
    \theta^\prime=\theta - \alpha \nabla_\theta \mathcal{L}_{CE}(\mathcal{D}_s; \theta),
\end{equation}
where $\alpha$ is the learning rate in the inner loop update. Then, we leverage a meta-learning step to enhance the generalization ability of the model by simulating the domain shift in the meta-test data $\mathcal{D}_t$. Specifically, in the meta-learning phase, the meta-objective $\mathcal{L}_{meta}$ is computed with the updated parameters $\theta^\prime$, but optimized towards the original parameters $\theta$. Intuitively, apart from learning the segmentation task on the meta-train data $\mathcal{D}_s$, this learning scheme teaches the model to learn how to adapt to the domain shift across {meta-training} data $\mathcal{D}_s$ and {meta-testing} data $\mathcal{D}_{s\rightarrow t}$. 

\textbf{Consistency Optimization.} While the meta-training scheme can simulate domain shifts and help the model adapt to these shifts, it primarily focuses on making the model robust under shifts within the source domain (as simulated with meta-source and meta-target domains). However, this approach alone is not sufficient to ensure that the model learns the representations that are invariant under different domains.
When training a model generalizable for unseen domains, it is crucial to guide the model in learning representations that remain consistent across multiple domains. This is important because, in the feature space, all the labeled data from source domains contribute to training a model that can generalize to previously unseen domains \cite{liDomainGeneralizationAdversarial2018}. If the data distributions across various domains within a given feature space remain dissimilar, it hinders the model's ability to effectively generalize to unseen domains. This is why we propose a consistency optimization method to force the model to learn the domain-invariant representations.

Specifically, consider that a collaborative BEV segmentation model $f( x_i^s; \theta)$ consists of an encoder $E$ and a segmentation decoder head $D_{seg}$, where $x_i^s$ is the input image and $\theta$ is the model parameters. The encoder $E$ will extract the perception features from different CAVs and fuse the features. The segmentation decoder head $D_{seg}$ will predict the segmentation mask.
We formulate the meta-train forward propagation on source domain images as:
\begin{equation}
    \hat{y}_i^s = f(x_i^s; \theta) = D_{seg}(E(x_i^s; \theta_E); \theta_D),
\end{equation}
where model parameters $\theta$ are decoupled into $\theta_E$ and $\theta_D$ for encoder and decoder, respectively.
Then, after computing the {meta-training} loss $\mathcal{L}_{CE}(x_i^s; \theta)$, we can obtain the updated model parameters $\theta^\prime$ by the inner loop update (Here, the updated model parameters $\theta^\prime$ will just be used in the {meta-testing} step, and will be released after current iteration, the real updated parameters will be computed later). Then, we leverage the updated model parameters $\theta^\prime$ to compute the {meta-testing} forward propagation prediction on target domain images $x^t_i\in\mathcal{D}_{s\rightarrow t}$ that are generated by our amplitude augmentation method:
\begin{equation}
    \hat{y}_i^t = f(x_i^t; \theta^\prime) = D_{seg}(E(x_i^t; \theta_E^\prime); \theta_D^\prime).
\end{equation}
\textcolor{r2color}{Next, we extract the latent representation from input images $x_i^s$ and target domain images $x_i^t$ in the meta-training and meta-testing steps,} respectively:
\begin{equation}
    z^s = E(x_i^s; \theta_E),~ z^t = E(x_i^t; \theta_E^\prime).
\end{equation}

In order to enhance the domain-invariant feature capture capability of the model, we design the consistency loss function that leverages the maximum mean discrepancy \cite{grettonKernelMethodTwoSampleProblem2006} to make the latent representations $z^s$ and $z^t$ more similar. We first map the features to a reproducing kernel Hilbert space (RKHS) $\mathcal{F}$, and then compute the distance between their mean in RKHS as follows:
\begin{equation}
    \mathcal{L}_{cons} = \left\| \frac{1}{n_s} \sum_{i=1}^{n_s} \phi(z_i^s) - \frac{1}{n_t} \sum_{i=1}^{n_t} \phi(z_i^t) \right\|^2_\mathcal{F},
\end{equation}
where $\phi$ is the kernel function that maps the distribution to RKHS, and $n_s$ and $n_t$ are the numbers of samples in the source and target domain, respectively. Typically, the Gaussian RBF kernel is used as the kernel function:
\begin{equation}
    k(x, y) = \exp\left(-\frac{\|x - y\|^2}{2\sigma^2}\right).
\end{equation}
Then, the consistency loss function is given by \cite{grettonKernelMethodTwoSampleProblem2006}:
\begin{equation}
    \label{eq:consistency_loss}
    \begin{split}
         \mathcal{L}_{cons}&= \frac{1}{n_s^2} \sum_{i,j=1}^{n_s} k(z_i^s, z_j^s)+ \frac{1}{n_t^2} \sum_{i,j=1}^{n_t} k(z_i^t, z_j^t) \\
        &  - \frac{2}{n_s n_t} \sum_{i,j=1}^{n_s, n_t} k(z_i^s, z_j^t).
    \end{split}
\end{equation}

\begin{algorithm}[t]
    \caption{Meta-consistency Training }
    \label{alg:meta}
    \begin{algorithmic}[1]
    \STATE \textbf{Input:} Meta-train data $\mathcal{D}_s$, Meta-test data $\mathcal{D}_t$, Learning rates $\alpha$, $\gamma$, Consistency weight $\beta$
    \STATE \textbf{Initialize:} Model parameters $\theta$
    \WHILE{not converged}
        \STATE \textbf{{Meta-Train:}}
        \STATE  Obtain the prediction using meta-train data $\mathcal{D}_s$: $\hat{y}^s_i=D_{seg}(E(x_i^s; \theta_E); \theta_D)$
        \STATE  Obtain the latent feature $z^s=E(x_i^s; \theta_E)$
        \STATE  Compute the cross entropy loss $\mathcal{L}_{CE}$ with prediction $\hat{y}^s_i$ and ground truth $y_i^s$
        \STATE  Update parameters $ \theta^\prime=\theta - \alpha \nabla_\theta \mathcal{L}_{CE}(\mathcal{D}_s; \theta)$
        \STATE \textbf{Meta-Test:}
        \STATE Obtain the prediction using meta-test data $\mathcal{D}_t$ with updated parameters:  $\hat{y}^t_i=D_{seg}(E(x_i^t; \theta_E^\prime); \theta_D^\prime)$
        \STATE Obtain the latent feature $z^t=E(x_i^t; \theta_E^\prime)$
        \STATE Compute the cross entropy loss $\mathcal{L}_{CE}$ with prediction $\hat{y}^t_i$ and ground truth $y_i^t$
        \STATE \textbf{Consistency Optimization:}
        \STATE  Compute consistency loss $\mathcal{L}_{cons}$ with two latent features $z^s$ and $z^t$ by Eq. (\ref{eq:consistency_loss})
        \STATE  Obtain the final updated parameters $\theta \leftarrow \hat{\theta}$ in this iteration by Eq. (\ref{eq:meta_update})
    \ENDWHILE
    \end{algorithmic}
\end{algorithm}

\textbf{Overall learning objective.} The overall {meta-learning} objective is composed of the cross entropy loss $\mathcal{L}_{CE}$ and the consistency loss $\mathcal{L}_{cons}$ as follows:
\begin{equation}
    \mathcal{L}_{meta} = \mathcal{L}_{CE}(x_i^t; \theta^\prime) + \beta \mathcal{L}_{cons}(x_i^s, x_i^t; \theta^\prime),
\end{equation}
where $\beta$ is the balancing weight on the consistency loss, while $\theta^\prime$ are the updated model parameter obtained from the inner loop update in Eq. (\ref{eq:inner_loop}). Then, the inner loop objective
and the {meta-learning} objective are optimized together with respect to the original parameters $\theta$ as follows:
\begin{equation}
    \label{eq:meta_update}
    \hat{\theta} = \theta -\gamma\nabla\left( \mathcal{L}_{CE}(x_i^s; \theta) + \mathcal{L}_{meta}(x_i^s, x_i^t; \theta^\prime)\right).
\end{equation}
The whole training procedure is summarized in Alg. \ref{alg:meta}. 

\subsection{Intra-system Domain Alignment During Inference}

\begin{figure}[t]
    \centering
     \includegraphics[width=1\linewidth]{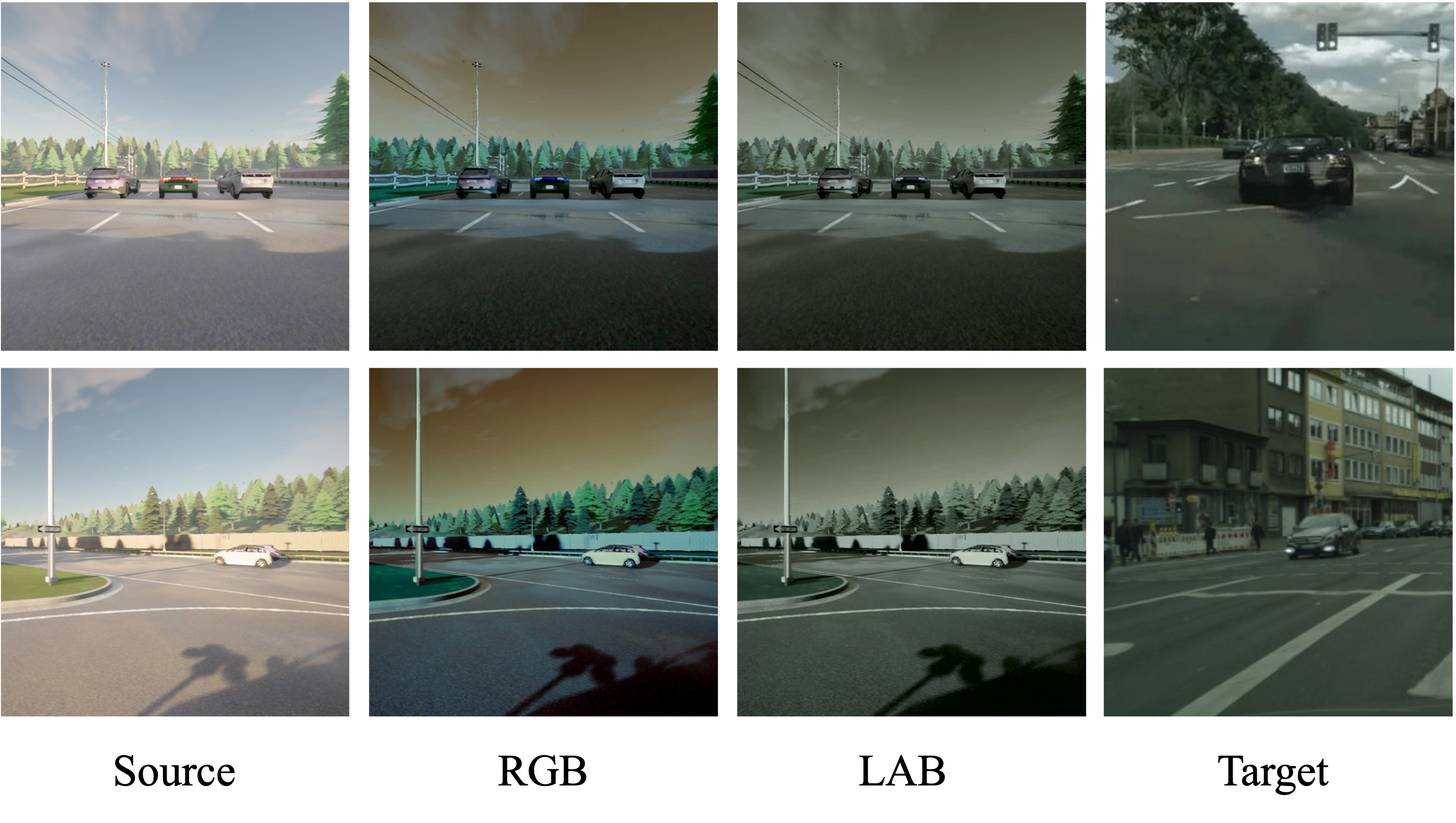}
  
    \caption{\textbf{Qualitative comparison} of image translation in the RGB and LAB color spaces.}
    \label{fig:image_translate}
\end{figure}

In collaborative perception for CAV driving, we observe that the ego \textcolor{r2color}{CAV} and other \textcolor{r2color}{CAVs} are situated at different environments, e.g., unbalanced lighting conditions: one \textcolor{r2color}{CAV} in the shade and the other in the open. Moreover, different types of car cameras can cause chromatic aberration. These phenomena can result in intra-system domain discrepancy, which can degrade the performance of a collaborative perception system, to some extent. In this subsection, we propose a simple yet effective method to align the intra-system domain shift during inference. This method can translate the image style by unifying the pixel distribution in color space. Inspired by \cite{reinhardColorTransferImages2001}, we employ image translation based on the LAB color space (LAB color space is another color model)  to minimize domain discrepancy. Similar to RGB, LAB color space is also a three-dimensional color space. The three dimensions of the LAB color space are $L^*$, $a^*$, and $b^*$, respectively, where $L^*$ represents the brightness of the color, and $a^*$ and $b^*$ represent the color components. 
This approach is motivated by two key observations. Firstly, the LAB color space has a broader gamut compared to the RGB color space. Secondly, as demonstrated in Fig. \ref{fig:image_translate}, images translated in the LAB color space more resemble the style of target domain images than those translated directly in the RGB color space. Thus, after considering a source domain RGB image $x^s$ and a target domain RGB image $x^t$, we convert the RGB image to the XYZ color space by:
\begin{equation}
    \left[\begin{array}{c}
    X \\
    Y \\
    Z
    \end{array}\right]=\left[\begin{array}{lll}
    0.4124 & 0.3575 & 0.1804 \\
    0.2126 & 0.7151 & 0.0721 \\
    0.0193 & 0.1191 & 0.9502
    \end{array}\right]\left[\begin{array}{l}
    R \\
    G \\
    B
    \end{array}\right].
\end{equation}
Then, we can obtain the LAB color space using the following equation:
\begin{equation}
    \begin{aligned}
    L^{\star} & =116 f\left(Y / Y_n\right)-16, \\
    a^{\star} & =500\left[f\left(X / X_n\right)-f\left(Y / Y_n\right)\right], \\
    b^{\star} & =200\left[f\left(Y / Y_n\right)-f\left(Z / Z_n\right)\right],
    \end{aligned}
\end{equation}
where $f(t)= t^{1 / 3}$, if $t>\left(\frac{6}{29}\right)^3$, otherwise $f(t)= \frac{1}{3}\left(\frac{29}{6}\right)^2 t+\frac{4}{29}$, while 
$X_n, Y_n, Z_n$ are the coordinate values of the reference white point. The LAB color space is a perceptually uniform color space, which means that a small change in the LAB color space corresponds to a small change in the perceived color. Therefore, \textcolor{r2color}{we can translate the image style by modifying the numerical values in the LAB color space}. After converting the RGB images to the LAB color space, denoted as $x^s_{LAB}$ and $x^t_{LAB}$, respectively, we first compute the mean and standard deviation of the LAB color channels of $x^s_{LAB}$ and $x^t_{LAB}$, denoted as $\mu^s_{LAB}$, $\sigma^s_{LAB}$, $\mu^t_{LAB}$, $\sigma^t_{LAB}$, respectively. Then, we can obtain the translated image $x^s_{LAB} \rightarrow x^t_{LAB}$ by:
\begin{equation}
    \hat{x}^s_{LAB} = \sigma^t_{LAB} / \sigma^s_{LAB} \cdot (x^s_{LAB} - \mu^s_{LAB}) + \mu^t_{LAB}.
\end{equation}
After obtaining the translated image $\hat{x}^s_{LAB}$, we can convert it back to the RGB color as $\hat{x}^s$.

In summary, in collaborative perception, the ego \textcolor{r2color}{CAV} first converts its perceived RGB images to the LAB color space and obtains $\mu^t_{LAB}$ and $\sigma^t_{LAB}$. Then, the ego \textcolor{r2color}{CAV} shares $\mu^t_{LAB}$ and $\sigma^t_{LAB}$ with other \textcolor{r2color}{CAVs}. Other \textcolor{r2color}{CAVs} can then use these parameters to translate their perceived RGB images to the target domain style. This process helps reduce the domain discrepancy among the collaborative \textcolor{r2color}{CAVs}. After that, other \textcolor{r2color}{CAVs} extract the intermediate features from the translated images and send these features to the ego vehicle. The ego \textcolor{r2color}{CAV} then fuses these features to predict the final segmentation mask.

\begin{figure}[t]
    \centering
     \includegraphics[width=1\linewidth]{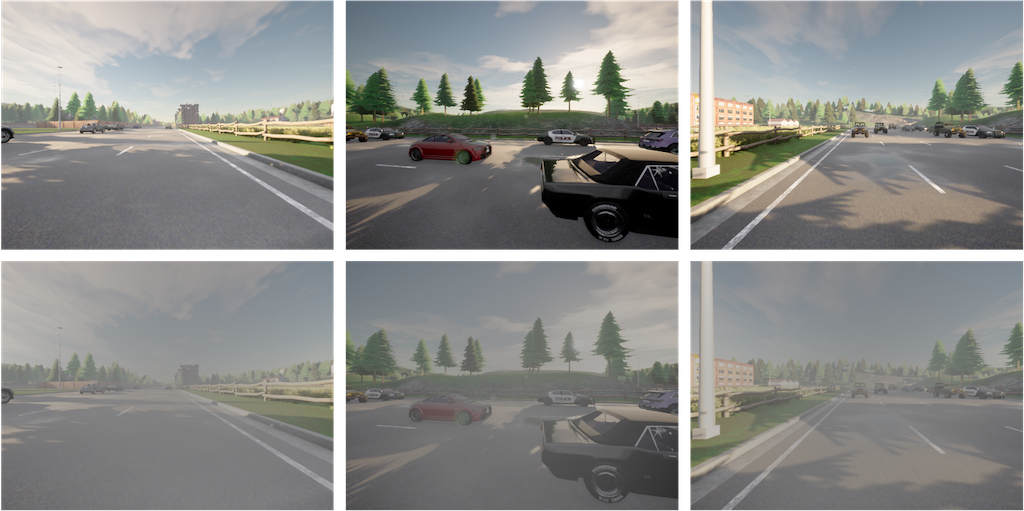}
  
    \caption{\textbf{Visualization of the foggy dataset}. The first row is the original image, and the second row is the synthesized foggy image.}
    \label{fig:visualization_fog}
\end{figure}
\begin{figure}[t]
    \centering
    \includegraphics[width=1\linewidth]{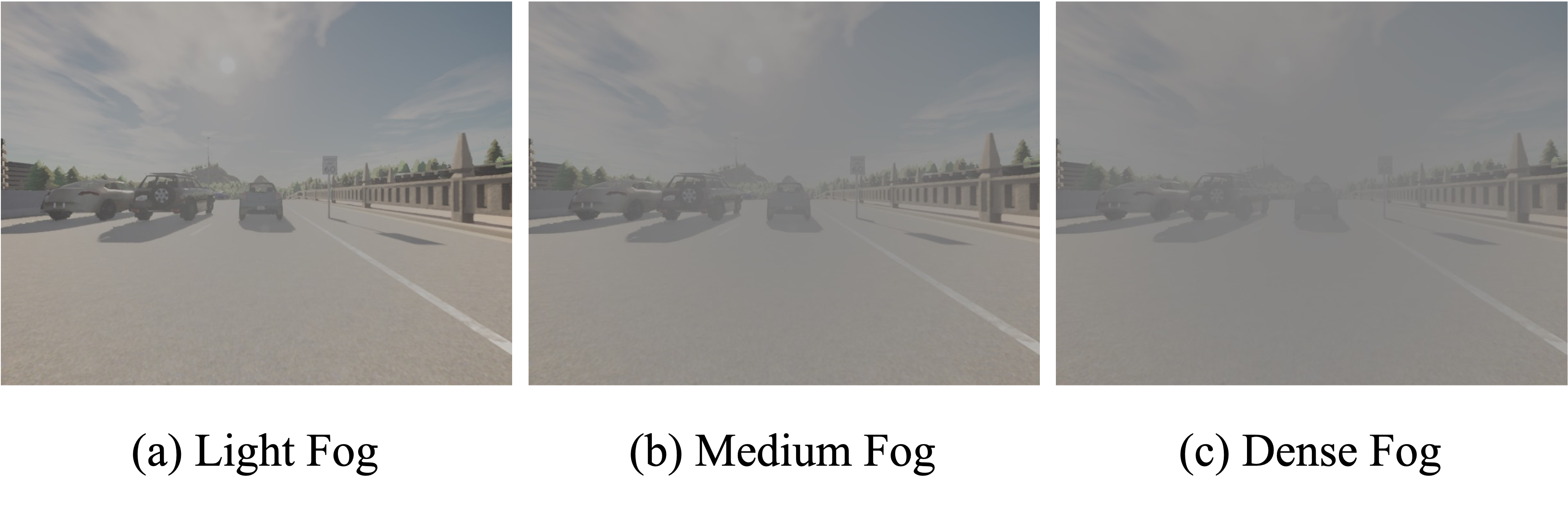}
    \caption{{Visualization of the fog with different densities.}}  
    \label{fig:visualization_fog_density}
    \vspace{-3mm}
\end{figure}

\begin{figure}[t]
    \centering
     \includegraphics[width=1\linewidth]{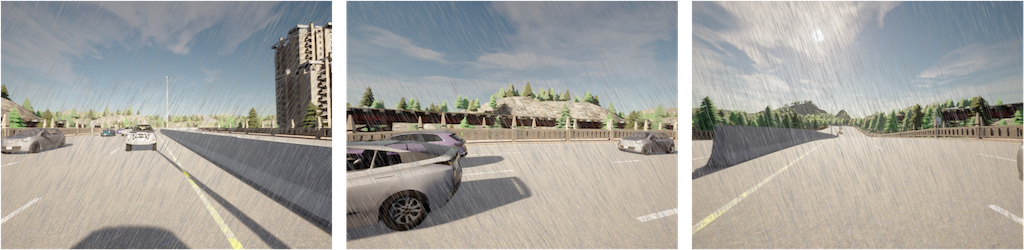}
  
    \caption{{Visualization of the rainy dataset. }}
    \label{fig:rain}
    \vspace{-3mm}
\end{figure}

\begin{figure}[t]
    \centering
     \includegraphics[width=1\linewidth]{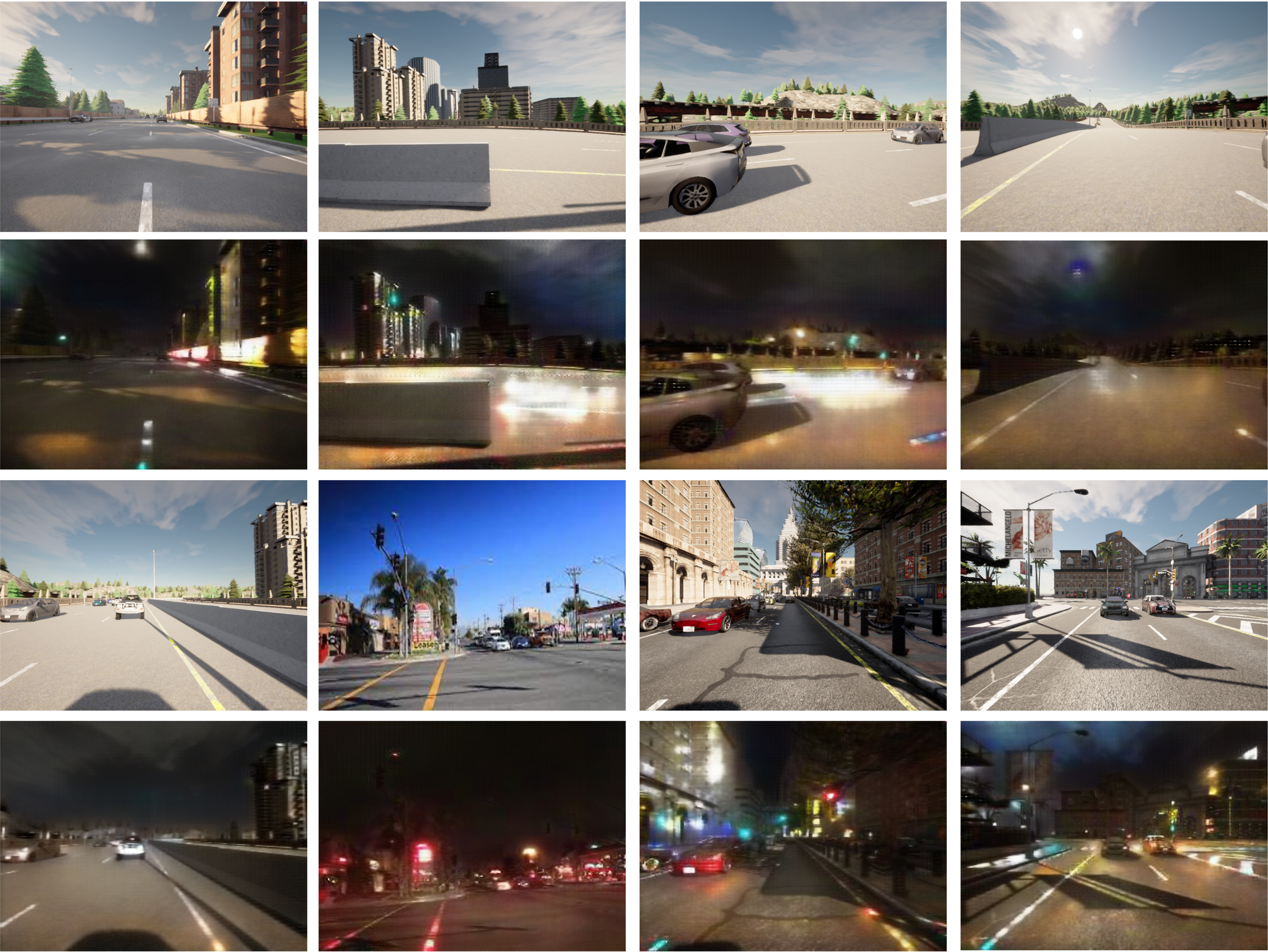}
  
    \caption{\textbf{Visualization of the night dataset}, the first row and the third row are the original images while the second row and the fourth row are the synthesized night images. }
    \label{fig:night}
    \vspace{-3mm}
\end{figure}

\section{Experiments}
\label{sec:experiments}

In this section, we conduct extensive experiments to evaluate the effectiveness of our proposed method. Specifically, we choose OPV2V as our fundamental dataset and synthesize four different datasets with different domain discrepancies. 
\textcolor{r2color}{We first introduce the datasets and metrics used in our experiments in Sec. \ref{sec:datasets}. Then, we present the implementation details of our method in Sec. \ref{sec:implementation_details}. Next, we compare our method with several state-of-the-art methods in Sec. \ref{sec:comparison_analysis}. Finally, we conduct ablation studies to evaluate the effectiveness of our method in Sec. \ref{sec:ablation_study}.}

\subsection{Datasets and Metrics}
\label{sec:datasets}

To assess the domain generalization performance of our proposed method, it is essential to evaluate it across various datasets exhibiting domain discrepancies. However, in the domain of collaborative BEV segmentation, there are no public available datasets suitable for this purpose. Consequently, we have taken the initiative to synthesize four distinct datasets, each characterized with unique domain discrepancies, drawing from the OPV2V dataset \cite{xuOPV2VOpenBenchmark2022} as the foundation. These datasets include variations,  {including} foggy weather, rainy weather, night conditions, and bright sunny scenarios. The specific attributes and details of these synthesized datasets are elaborated below.

\textbf{OPV2V.} In our study, we leverage the OPV2V dataset \cite{xuOPV2VOpenBenchmark2022}, a comprehensive dataset curated for integrated perception using V2V communications. This dataset, amassed through the CARLA simulator \cite{dosovitskiyCARLAOpenUrban2017} and OpenCDA \cite{xuOpenCDAOpenCooperative2021}, comprises 11,464 frames of LiDAR point clouds and images, with each frame featuring the minimum of 2 and the maximum of 7 interconnected vehicles. It encompasses 73 distinct scenarios, each averaging a duration of 25 seconds.

\textbf{Sunny.} The initial task in our setting is driving on a sunny day. As the OPV2V dataset is collected in the daytime, so we sample a portion of OPV2V as our sunny dataset.

\textbf{Fog.} The second task is driving on a foggy day. We synthesize a foggy-weather dataset with different densities of fog (e.g., dense fog and light fog) as shown in Fig. \ref{fig:visualization_fog} and \ref{fig:visualization_fog_density}, respectively. The foggy weather is generated from the atmospheric scattering model \cite{narasimhanVisionAtmosphere2008}.

\textbf{Rain.} In addition, we synthesize a rainy-weather dataset with different densities of rain (e.g., heavy rain and light rain) as shown in Fig. \ref{fig:rain}. Specifically, the dataset is generated by utilizing RainMix, which is a novel data augmentation method. In order to generate the rainy dataset, we first sample a rain map from a public dataset of real rain streaks \cite{gargPhotorealisticRenderingRain2006}. We then apply random transformations to the rain map using the RainMix technique. These transformations include rotation, zooming, translation, and shearing, which are randomly sampled and combined. Lastly, the rain maps after transformation are merged with the original source domain images.

\textbf{Night.} Finally, as shown in Fig. \ref{fig:night}, we synthesize a night dataset by leveraging a day-to-night generative adversarial network (GAN) \cite{zhuUnpairedImageImageTranslation2017}, which can render the daytime images to night images. \textcolor{cvprblue}{The four datasets under different weather conditions contain 13,920 images in total, and each dataset contains 3,480 images.}

For performance assessment, we utilize the Intersection of Union (IoU) metric to contrast the predicted map with the actual map-view labels.
\begin{equation}
    \text{IoU} = \frac{|BEV_1 \cap BEV_0|}{|BEV_1 \cup BEV_0|},
\end{equation}
where $BEV_0$ represents the ground truth BEV map, $BEV_1$ represents the predicted BEV map. The IoU metric is calculated for each class and averaged across all classes to obtain the final IoU score. 

\begin{table*}[t]
    \caption{\textbf{Comparison results and evaluation} of the domain generalization performance with different domains.}
    \label{fig:comparing_results}
    \normalsize
    \begin{tabular*}{\hsize}{@{}@{\extracolsep{\fill}}p{3cm}|cccc|cccc@{}}
    \hline
    
    Domain Type  & \multicolumn{4}{c|}{Sunny}        & \multicolumn{4}{c}{Fog}           \\ \hline
    AP@IoU(\%)   & Vehicle & Road  & Lane  & Average & Vehicle & Road  & Lane  & Average \\ \hline
    AttFuse \cite{xuOPV2VOpenBenchmark2022}  & 51.92   & 36.89 & 22.96 &  37.27       & 40.20   & 11.54 & 0.04  &  18.58       \\
    V2VAM \cite{liLearningVehicletoVehicleCooperative2023} &48.85&35.14&20.08&34.69&41.83&21.56&8.04&23.81\\
    F-Cooper \cite{chenFCooperFeatureBased2019} &42.27&32.34&25.01&33.17&32.45&12.83&9.04&18.10\\
    Where2Comm \cite{huWhere2commCommunicationefficientCollaborative2024} &44.13&30.50&24.89&33.17&26.70&5.91&9.00&13.87\\
    V2X-ViT \cite{xuV2XViTVehicletoEverythingCooperative2022a}&37.86&36.35&24.91&33.04&25.79&23.13&10.19&19.70\\
    DiscoNet \cite{liLearningDistilledCollaboration2021} & 39.83   & \textbf{53.19} & 37.37 &   43.46      & 33.74   & 33.68 & 23.61 &   30.34      \\
    V2VNet \cite{wangV2VNetVehicletoVehicleCommunication2020a}  & 41.33   & 51.32 & 37.21 &  43.29       & 35.05   & 36.06 & 22.10 &    31.07     \\ 
    CoBEVT \cite{xuCoBEVTCooperativeBird2023}     & 52.16   & 46.58 & 38.07 &  45.60       & 25.43   & 22.69 & 7.16 &   18.42      \\ 
    \hline
    Ours     & \textbf{55.05}   & 50.84 & \textbf{39.48} &  \textbf{48.46}       & \textbf{42.01}   & \textbf{41.20} & \textbf{28.33} &   \textbf{37.18}      \\ \hline\hline
    Domain  Type & \multicolumn{4}{c|}{Rain}         & \multicolumn{4}{c}{Night}         \\ \hline
    AP@IoU(\%)   & Vehicle & Road  & Lane  & Average & Vehicle & Road  & Lane  & Average \\ \hline
    AttFuse \cite{xuOPV2VOpenBenchmark2022} & 47.93   & 29.15 & 0.14  &    25.74     &22.58         &9.68       &2.22       &11.49         \\
    V2VAM \cite{liLearningVehicletoVehicleCooperative2023} &46.62&26.98&18.90&30.83&13.11&11.54&3.87&9.51\\
    
    F-Cooper \cite{chenFCooperFeatureBased2019} &42.50&26.32&18.52&29.11&14.59&9.21&5.27&9.66\\
    Where2Comm \cite{huWhere2commCommunicationefficientCollaborative2024} &40.26&28.31&24.26&30.94&11.78&11.91&7.99&10.56\\
    V2X-ViT\cite{xuV2XViTVehicletoEverythingCooperative2022a} &38.10&26.05&16.51&26.89&15.21&11.27&6.97&11.15\\
    DiscoNet \cite{liLearningDistilledCollaboration2021} & 36.83   & \textbf{45.48} & 29.96 &    37.42     &22.62         &5.41       &9.42       &  12.48       \\
    V2VNet \cite{wangV2VNetVehicletoVehicleCommunication2020a}  & 41.38   & 43.47 & 29.72 &    38.19     &22.43         &14.52       &4.73       &    13.89     \\ 
    CoBEVT \cite{xuCoBEVTCooperativeBird2023}     & \textbf{50.67}   & 29.33 & 28.24 &  36.08       &5.53    & 20.16 & 2.15&   9.28      \\ 
    
    \hline
    Ours     & 49.77   & 36.52 & \textbf{30.80} &    \textbf{39.03}     & \textbf{25.78}         & \textbf{21.61}      &  \textbf{12.73}     &    \textbf{20.04}     \\ \hline
    \end{tabular*}
    \vspace{-3mm}
\end{table*}

\subsection{Implementation Details}
\label{sec:implementation_details}

Our model is built on PyTorch and trained on two RTX4090 GPUs utilizing the AdamW optimizer. In training process, the initial learning rate is set to $2\times 10^{-4}$ and decays by an exponential factor of $1\times 10^{-2}$ with a cosine annealing learning rate scheduler.
During meta-consistency training, the meta-step size is set to $1\times 10^{-3}$. The hyperparameter $\alpha$ is empirically set to 0.01 to avoid artifacts on the transformed images. The balance weight $\beta$ of consistency loss is set to 0.1.
We set the communication range of CAVs to 70m, and all the CAVs outside the communication range from the ego CAV are ignored.
In addition, we first scale the camera images into $512\times 512$ pixels, then we employ CVT \cite{zhouCrossviewTransformersRealtime2022a} as our backbone to extract the features of each CAV, we employ FuseBEVT \cite{xuCoBEVTCooperativeBird2023} to fuse the features from different CAVs, and after that we employ a six layer convolutional neural network (CNN) as the decoder to generate the final BEV segmentation map.

\subsection{Comparison Analysis}
\label{sec:comparison_analysis}

\textbf{Experimental Setting.} 
In our experiment, we first train all the models in the original OPV2V dataset for 90 epochs with batch size equal to 1, then we continue to train the model by our proposed meta-consistency training paradigm for 20 epochs. After that, we evaluate the performance on the synthesized datasets. 
In assessing the performance of our BEV segmentation rendered by our proposed approach, we compare it with several methods, including AttFuse \cite{xuOPV2VOpenBenchmark2022}, V2VAM \cite{liLearningVehicletoVehicleCooperative2023}, F-Cooper \cite{chenFCooperFeatureBased2019}, Where2Comm \cite{huWhere2commCommunicationefficientCollaborative2024}, V2X-ViT \cite{xuV2XViTVehicletoEverythingCooperative2022a}, DiscoNet \cite{liLearningDistilledCollaboration2021}, V2VNet \cite{wangV2VNetVehicletoVehicleCommunication2020a}, and CoBEVT \cite{xuCoBEVTCooperativeBird2023}. 
For the implementation of these methods, we follow the original papers and use the open code provided by the authors and establish them in the collaborative BEV segmentation setting with the same decoder.

\textbf{Comparison Results.} The results of our experiments are summarized in Table \ref{fig:comparing_results}, showcasing the domain generalization performance across different environmental conditions. It is evident from the table that our proposed method significantly outperforms the state-of-the-art techniques across all {test domains}, affirming the robustness and efficacy of our approach. In the Sunny domain, our method achieves the highest average precision (AP) of 46.96\%, a substantial improvement over the next best performance of 45.60\% by CoBEVT. Similarly, in the Fog domain, our method again tops the list with an AP of 37.18\%, demonstrating a notable gain over 31.07\% achieved by V2VNet. Moreover, in challenging Rain and Night domains, our approach continues to exhibit superior performance. In the Rain domain, we achieve an AP of 39.03\%, outclassing the 38.19\% by V2VNet, and in the Night domain, our method significantly surpasses others with an AP of 20.04\%, with the next best being 13.89\% by V2VNet. The comparative analysis elucidates the benefit of employing a unified domain generalization framework for training and inference in CP. Our method's ability to generalize across various domains validates the efficacy of the employed amplitude augmentation and meta-consistency training scheme. The results also underline the importance of the intra-system domain alignment during the inference phase, significantly reducing the domain discrepancy among CAVs and consequently enhancing the collaborative perception performance across different environmental conditions.

\begin{figure*}[t]
    \centering
    \includegraphics[width=1\textwidth]{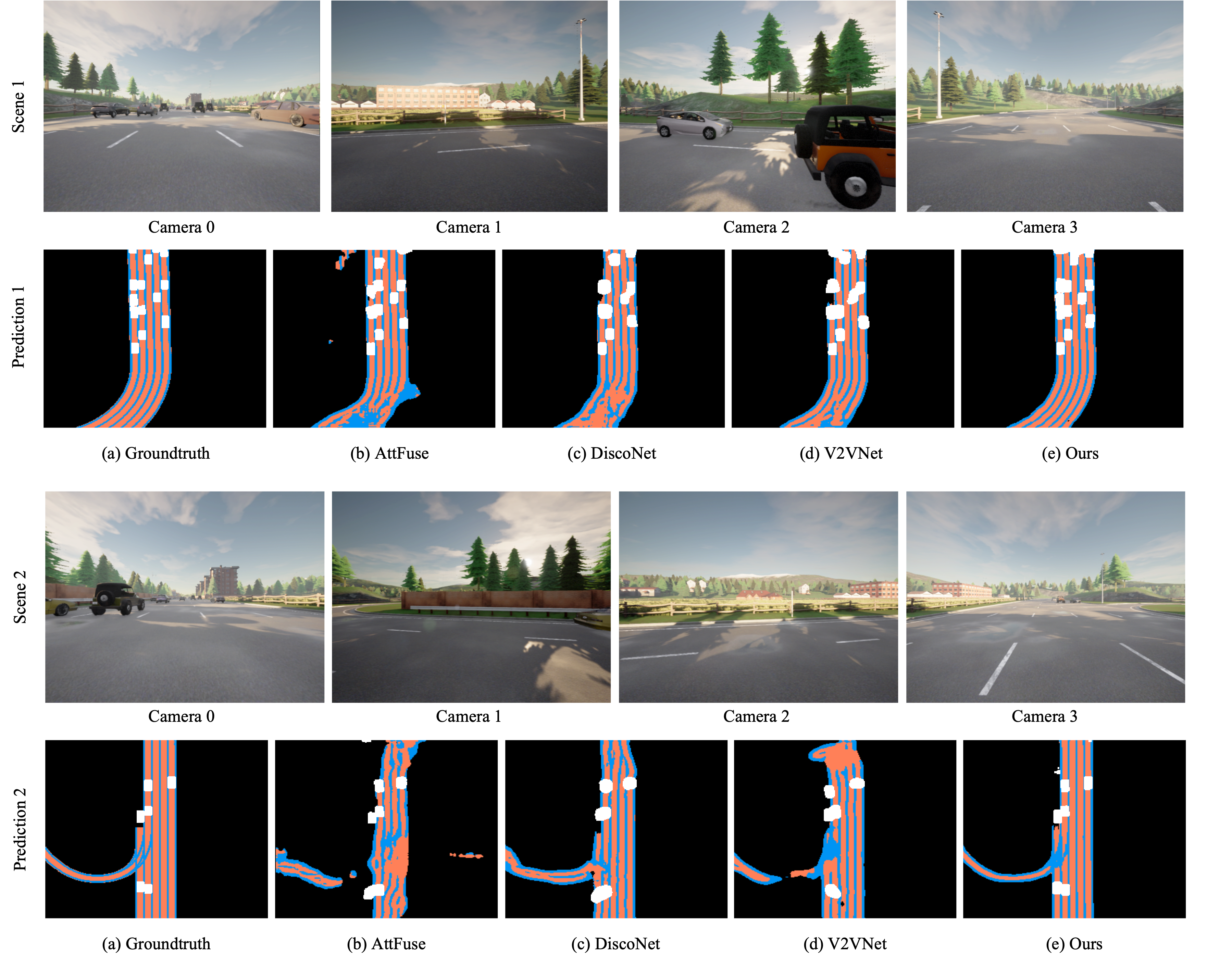}
    \caption{\textbf{Visualization of the BEV segmentation results of different methods}. In the first row, The four images are from ego vehicle's cameras from four different angles. In the second row, the five BEV maps are ground truth and generated from different method: column (a) is the ground truth, (b) is generated from AttFuse, (c) is from DiscoNet, (d) is from V2VNet, and (e) is predicted by our method. The third row and forth row are for another scene. Compared with other methods, our method shows robust performance in different scenes.}
    \label{fig:qualitative_visualization}
    \vspace{-3mm}
\end{figure*}

\textbf{Qualitative Analysis.} In order to provide a qualitative comparison across different methods, we visualize some BEV segmentation results of different methods (including AttFuse, DiscoNet, V2VNet) in Fig. \ref{fig:qualitative_visualization}. As we observe in the BEV prediction maps, our model yields the perception results that stand out in terms of both comprehensiveness and accuracy when compared with other methods. In both scenes, AttFuse exhibits significant omissions in road surface and lane. DiscoNet and V2VNet more or less fail to segment the complete road and lane. As for the vehicle class, these three compared methods occasionally miss segments and display ambiguous boundaries. In contrast, as seen in Fig. \ref{fig:qualitative_visualization} (e), our method can segment the road, lane, and vehicle class more accurately and comprehensively. These results can show the superiority of our method in the CP task.

\subsection{Ablation Study}
\label{sec:ablation_study}

\begin{figure*}[t]
    \centering
     \includegraphics[width=1\textwidth]{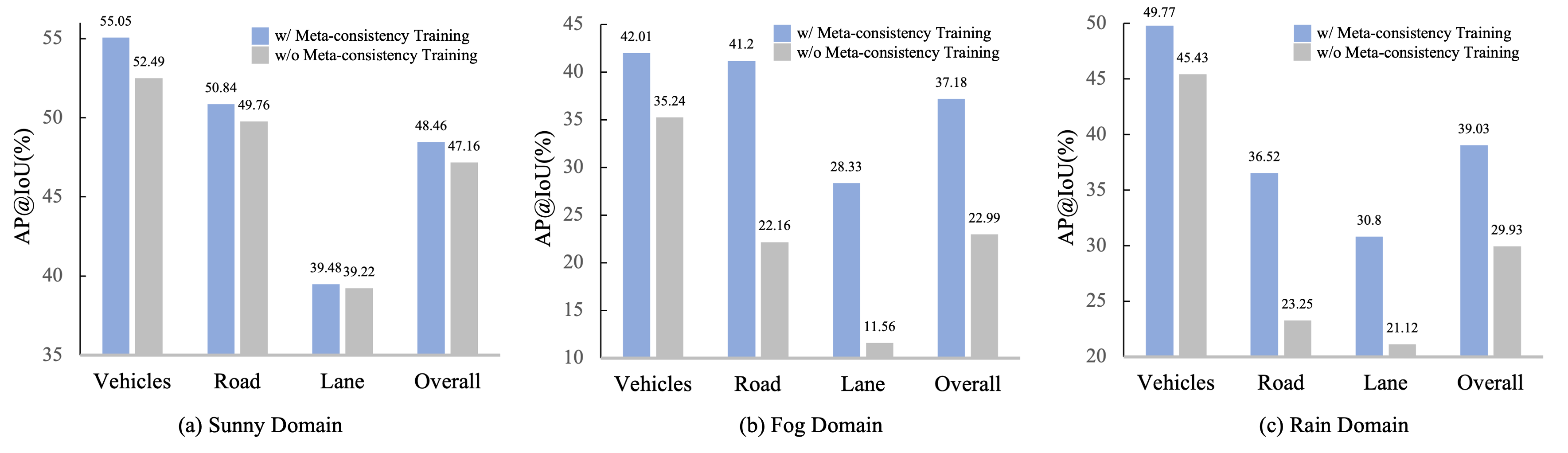}
  
    \caption{\textbf{Effect of the meta-consistency training.} ``w/'' means \textit{with meta-consistency training}, ``w/o'' means \textit{without meta-consistency training}.}
    \label{fig:effect_of_meta_consistency}
    \vspace{-3mm}
\end{figure*}

\begin{figure*}[t]
    \centering
    \includegraphics[width=1\textwidth]{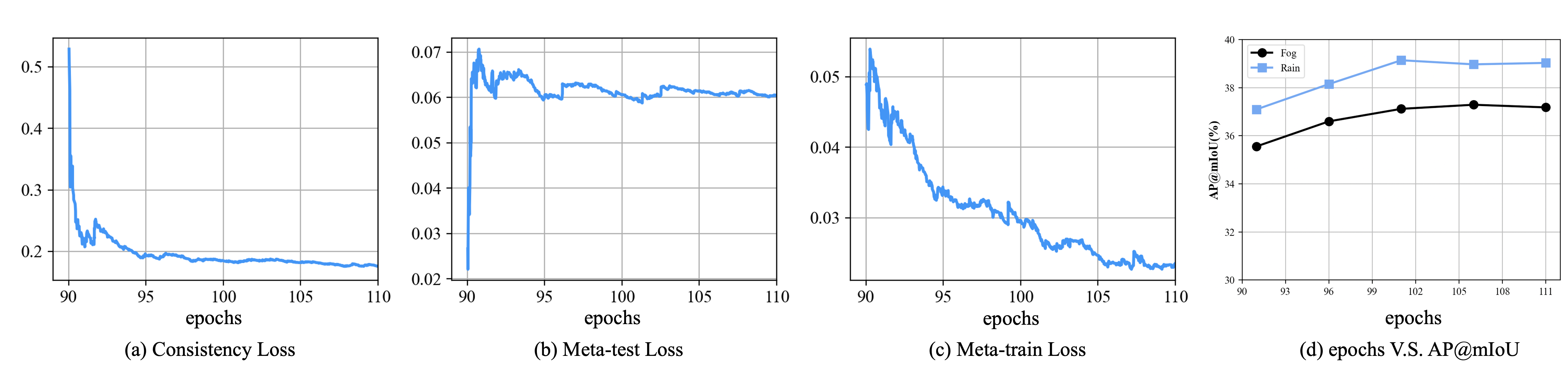}
    \caption{Visualization of the consistency loss $\mathcal{L}_{cons}$, meta-test loss, and meta-train loss. \textcolor{cvprblue}{Subfigure (d) visualizes AP@IoU tested on the foggy and rainy domains with different epochs.}}
    \label{fig:consistency_loss}
    \vspace{-3mm}
\end{figure*}

\textbf{Effect of Amplitude Augmentation.}
In order to evaluate the effectiveness of the proposed amplitude augmentation, we set the sunny dataset to the unseen domain and test its performance. Then, we conduct an experiment depicted in Table \ref{fig:effect_of_amplitude_augmentation}. We observe that by simply removing the amplitude augmentation, the performance degrades. The IoU accuracy of vehicles, road, and lane are decreased by 9.51\%, 5.58\%, and 3.51\%, respectively. 
In addition, we conduct comparative experiments with conventional augmentation methods, RainMix \cite{gargPhotorealisticRenderingRain2006} and Color Augmentation (ColorAug) \cite{otaloraStainingInvariantFeatures2019}, to show the efficiency of AmpAug. In the experiment, we replace the AmpAug module with others, respectively. From Table \ref{fig:effect_of_amplitude_augmentation}, we observe that our AmpAug can achieve better performance than the other two methods, further demonstrating the effectiveness of our AmpAug.

\begin{figure}[t]
    \centering
    \includegraphics[width=.9\linewidth]{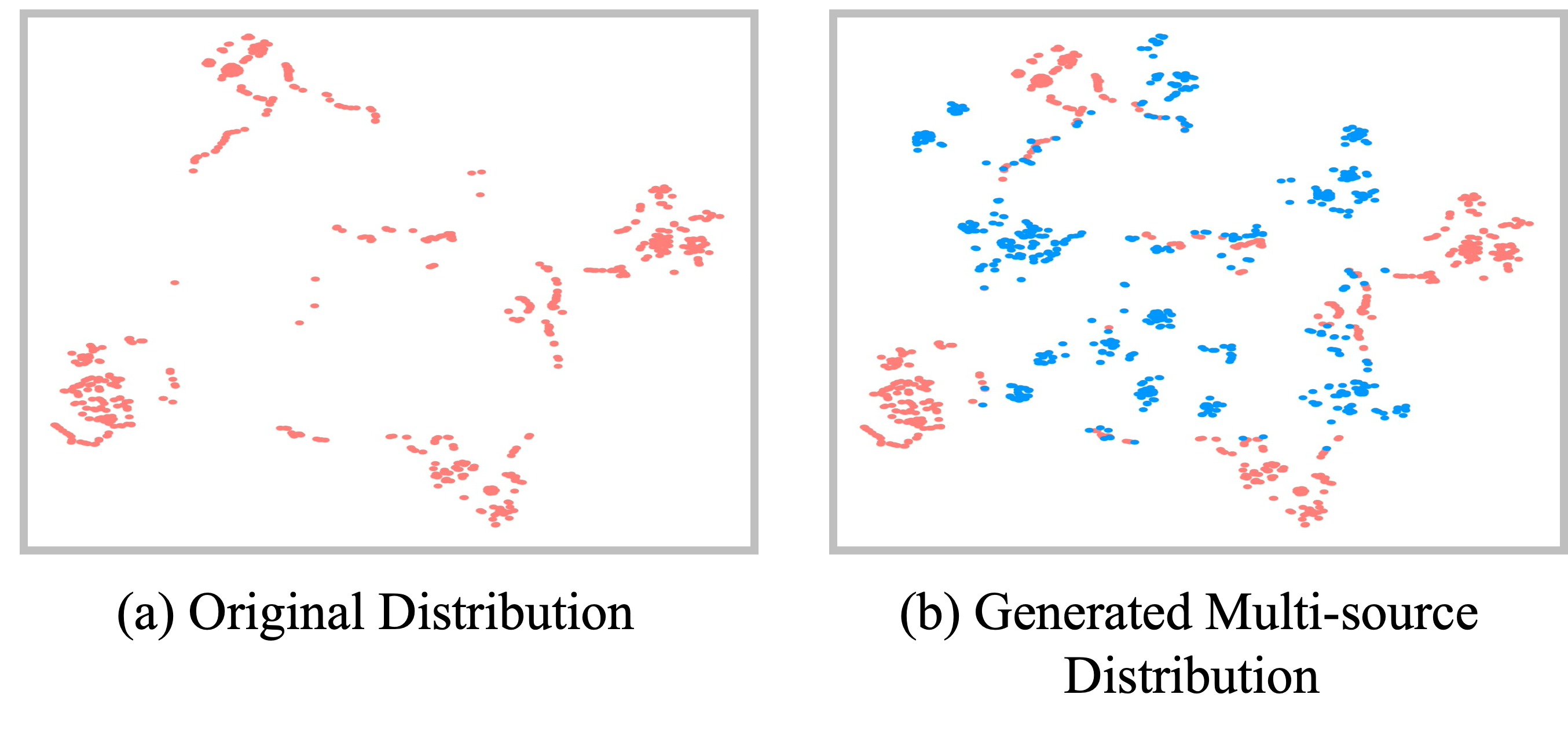}
    \caption{\textbf{Visualization of the t-SNE embedding} for the original dataset (pink points) and the corresponding transformed images by AmpAug dataset (blue points). }
    \label{fig:tsne}
    \vspace{-3mm}
\end{figure}

To further capture the effect of AmpAug and show that AmpAug can increase the diversity of the source domain. We use t-SNE\footnote{\label{t-SNE}t-SNE: t-distributed stochastic neighbor embedding is a statistical method for visualizing high-dimensional data by giving each data point a location in a two or three-dimensional map.} \cite{van2008visualizing} to visualize the distribution of the original images and the generated images in Fig. \ref{fig:tsne}. Specifically, the pink points denote the local data, the other blue points denote the transformed data that are generated with AmpAug from our constructed dataset $\mathcal{D}_t$ proposed in Sec. \ref{sec:amplitude_augmentation}. From the observation, we can see that the distribution of the transformed images is more diverse than the original images. This indicates that AmpAug can increase the diversity of the source domain by augmenting the low-level distribution of the images, thereby making the model see more low-level variations. This can prevent the model from overfitting the source domain and help the model learn the domain-invariant features. 
\begin{table}[t]
    \centering
    \caption{\textcolor{r2color}{\textbf{Ablation Study and Comparative Results of AmpAug}. ``w/'' means \textit{with AmpAug}, ``w/o'' means \textit{without AmpAug.}}}
    \label{fig:effect_of_amplitude_augmentation}
    \resizebox{1\columnwidth}{!}{
    \begin{tabular}{c|cccc}
    \hline
    Methods & Vehicles & Road  & Lane & Average \\ \hline\hline
    w/o AmpAug       & 45.54   & 45.26 & 35.97 & 42.26 \\
    \textcolor{cvprblue}{RainMix \cite{gargPhotorealisticRenderingRain2006}} &47.74 & 46.56 & 35.88 &  43.39 \\
    \textcolor{cvprblue}{ColorAug \cite{otaloraStainingInvariantFeatures2019}} & 47.55 & 48.26 & 35.57 &  43.79 \\
    w/ AmpAug        & \textbf{55.05}   & \textbf{50.84} & \textbf{39.48} & \textbf{48.46}\\ \hline
    \end{tabular}}
    \vspace{-3mm}
\end{table}

\begin{table}[t]
    \centering
    \caption{\textbf{Ablation Study Results of the Intra-System Domain Alignment}. ``w/'' means \textit{with intra-system domain alignment}, ``w/o'' means \textit{without intra-system domain alignment.} }
    \normalsize
    \resizebox{1\columnwidth}{!}{
        \begin{tabular}{cc|cccc}
        \hline
        \multicolumn{2}{c|}{Methods}                                  & Vehicles & Road           & Lane       & Average        \\ \hline\hline
        \multicolumn{1}{c}{\multirow{2}{*}{AttFuse \cite{xuOPV2VOpenBenchmark2022}}}     & w/o  & 51.92 & 36.89         & 22.96               & 37.27       \\
        \multicolumn{1}{c}{}                               & w/   &\textbf{53.16} &\textbf{41.09  }        &\textbf{25.60}        & \textbf{39.95}      \\ \hline
        \multicolumn{1}{c}{\multirow{2}{*}{DiscoNet \cite{liLearningDistilledCollaboration2021}}}     & w/o  &39.83       &53.19               &37.37      & 43.46      \\
        \multicolumn{1}{c}{}                               & w/   &\textbf{41.94}  & \textbf{53.23 }        & \textbf{37.80}& \textbf{44.32}              \\ \hline
        \multicolumn{1}{c}{\multirow{2}{*}{V2VNet \cite{wangV2VNetVehicletoVehicleCommunication2020a}}}        & w/o  & 41.33 &         51.32&37.21          & 43.29              \\
        \multicolumn{1}{c}{}                               & w/   &  \textbf{44.85}&          \textbf{53.49}  &\textbf{39.13}    &\textbf{45.82}      \\ \hline
        \multicolumn{1}{c}{\multirow{2}{*}{Ours}} & w/o  & 54.09 & 49.13          & 37.66          & 46.96        \\
        \multicolumn{1}{c}{}                               & w/  & \textbf{55.05} & \textbf{50.84} & \textbf{39.48} & \textbf{48.46} \\ \hline
        \end{tabular}
    }
    \label{DA-ablation}
    \vspace{-3mm}
\end{table}

\textbf{Effect of Meta-consistency Training.}
In order to evaluate the effectiveness of the proposed meta-consistency training, we conduct an experiment depicted in Fig. \ref{fig:effect_of_meta_consistency}. We observe that the meta-consistency training does improve the performance of the model: the IoU accuracy of vehicles, road, and lane are improved by 2.56\%, 1.08\% and 0.26\%, respectively. Without our method, the model could not learn the domain-invariant features, which leads to performance degradation. Notably, the performance of the vehicle class is more sensitive to the domain discrepancy than the other classes, because vehicles are more dynamic than the other classes. The results demonstrate that our meta-consistency training can improve the performance of CP and make the model more robust to the domain shift.

To further analyze the effectiveness of the meta-consistency training, we visualize the consistency loss, denoted as $\mathcal{L}_{cons}$ in Fig. \ref{fig:consistency_loss}. It can be observed that despite the consistency loss $\mathcal{L}_{cons}$ has some fluctuations, it shows a decreasing trend during the training process. This suggests that the consistency loss can make the latent representations more similar, thereby aiding the model to learn the domain-invariant features. In addition, \textcolor{r2color}{the meta-testing loss and meta-training loss also show a decreasing trend. The meta-testing loss has a large fluctuation at the beginning of the training, but it gradually levels off around 0.06}. In addition, we also evaluate mIoU in the test set in fog and rain with different epochs, as shown in Fig. \ref{fig:consistency_loss}(d), we observe that the AP levels off when the number of epochs is around 100, which reflects the convergence of the loss.

\textbf{Effect of Intra-system Domain Alignment.}
In order to evaluate the effect and the university of our intra-system domain alignment mechanism, we conduct an experiment with different collaborative perception frameworks, including AttFuse \cite{xuOPV2VOpenBenchmark2022}, V2VNet \cite{wangV2VNetVehicletoVehicleCommunication2020a}, and DiscoNet \cite{liLearningDistilledCollaboration2021}. Before evaluating these different collaborative perception frameworks, we first implement our method on each framework.
The results are shown in Table \ref{DA-ablation}. We observe that our intra-system domain alignment mechanism can generally improve the performance of the baselines, especially for the vehicle class. Specifically, for AttFuse, the IoU accuracy of vehicles, road, and lane are improved by 1.24\%, 4.23\%, and 2.64\%, respectively. For V2VNet, the IoU accuracy of vehicles, road, and lane are improved by 3.52\%, 2.17\%, and 1.92\%, respectively. In addition, as for our method, we simply remove this component and evaluate it, we can see that the performance of the model will be degraded. These results can demonstrate the effect of our intra-system domain alignment mechanism that can generally improve the performance of our CP framework by reducing the distribution heterogeneity among the data in different vehicles.

\textcolor{cvprblue}{\textbf{Collective Ablation Study.} In this section, we further analyze the collective impact of these techniques on the performance of our overall framework and provide a deeper understanding of their respective contributions and synergistic effects within the CP framework. As shown in Tab. \ref{r1q8}, we conduct an ablation study of our proposed method. We compare the performance of our method with the vanilla method, which removes the three techniques in our method, including AmpAug, meta-consistency training, and intra-system domain alignment. The results show that our method outperforms the vanilla method in most cases, which indicates that each component contributes to the overall performance.}

\begin{table}[t]
    \centering
    \caption{ \textcolor{cvprblue}{\textbf{Collective Ablation Study Results of the Whole Techniques.}. \textit{`Vanilla'} represents that we remove the three techniques in our method, including AmpAug, meta-consistency training, and intra-system domain alignment.}}
    \label{r1q8}
    \normalsize  
    \resizebox{1\columnwidth}{!}{
    \begin{tabular}{c|c|cccc}
    \hline
    Method                   & Domain & Vehicles & Road  & Lane  & Average \\ \hline\hline
    \multirow{4}{*}{Vanilla} 
    & Sunny  & 46.67    & 47.33 & 36.66 & 43.55   \\
                             & Fog    & 44.23    & 35.37 & 26.88 & 35.56   \\
                             & Rain   & 42.35    & 33.10 & 29.34 & 34.93   \\
                             & Night  & 20.15    & 15.61 & 9.57  & 15.11   \\ 
                             \hline
    \multirow{4}{*}{Ours}    
    & Sunny  & 55.05    & 50.84 & 39.48 & 46.96   \\
                             & Fog    & 42.01    & 41.20 & 28.33 & 37.18   \\
                             & Rain   & 49.77    & 36.52 & 30.80 & 39.03   \\
                             & Night  & 25.78    & 21.61 & 12.73 & 20.04   \\ 
                             \hline
    \end{tabular}}
    \vspace{-3mm}
    \end{table}

\section{Limitations and Failure Cases}

\textcolor{r2color}{Our work introduces a novel framework to tackle the full-scene domain generalization problem in multi-agent collaborative BEV segmentation. While domain generalization is inherently challenging, and no existing works, including ours, can completely solve it for all scenarios, and thus, we acknowledge certain limitations and failure cases that require further investigation. For instance, although our method significantly mitigates domain shift, it struggles when the shift is substantial or when the target domain starkly differs from the source domain. A specific example is the performance discrepancy between day and night  scenarios. Despite the effectiveness of our techniques, the model's performance at night remains inferior to daytime results. We are committed to refining our approach and addressing these challenges in future work.}

\section{Conclusion}
\label{sec:conclusion}

In this paper, we have developed a novel framework to tackle the domain generalization with collaborative perception systems for connected and autonomous driving. To achieve this goal, we first propose an amplitude augmentation method to transform data from original source domain to various target domains in collaboration with our constructed dataset. Then, we design a meta-consistency training scheme based on meta-learning paradigm, which can simulate domain shifts, guiding the model to learn how to learn from different domains, and the meta-consistency loss can force the model to learn the domain-invariant features, thereby enhancing the ability to generalize. Finally, we  leverage the intra-system domain alignment to minimize the domain gap among the data perceived by different CAVs prior to inference. Comprehensive experiments have demonstrated the superiority of our framework compared with the existing state-of-the-art methods.

\bibliographystyle{IEEEtran}

{\small
\bibliography{MyLibrary2, ref}}

\vfill

\end{document}